\documentclass[11pt,a4paper]{article}
\usepackage[hyperref]{emnlp2018}
\usepackage{times}
\usepackage{latexsym}

\usepackage{url}

\aclfinalcopy 

\usepackage{amsmath}
\usepackage{bm}
\usepackage{color}
\usepackage{multirow}
\usepackage{subfig}
\usepackage{graphicx, wrapfig}

\usepackage{balance}

\DeclareMathOperator*{\argmax}{arg\,max}

\usepackage{xspace}

\newcommand{\eg}{\emph{e.g.}\xspace}
\newcommand{\ie}{\emph{i.e.}\xspace}

\usepackage{CJK}

\title{Learning to Jointly Translate and Predict Dropped Pronouns with a Shared Reconstruction Mechanism}

\author{Longyue Wang \\Tencent AI Lab\\{\tt vinnylywang@tencent.com} \And
  Zhaopeng Tu\thanks{~~~Zhaopeng Tu is the corresponding author of the paper. This work was conducted when Longyue Wang was studying and Qun Liu was working at the ADAPT Centre in the School of Computing at Dublin City University.} \\Tencent AI Lab\\{\tt zptu@tencent.com} \AND
  Andy Way \\ Dublin City University \\{\tt andy.way@adaptcentre.ie} \And
  Qun Liu\\Huawei Noah's Ark Lab\\{\tt qun.liu@huawei.com}}

\date{}

\begin{document}

\maketitle

\begin{abstract}
Pronouns are frequently omitted in pro-drop languages, such as Chinese, generally leading to significant challenges with respect to the production of complete translations. Recently,~\newcite{Wang:2018:AAAI} proposed a novel reconstruction-based approach to alleviating dropped pronoun (DP) translation problems for neural machine translation models. 
In this work, we improve the original model from two perspectives. 
First, we employ a shared reconstructor to better exploit encoder and decoder representations.
Second, we jointly learn to translate and predict DPs in an end-to-end manner, to avoid the errors propagated from an external DP prediction model.
Experimental results show that our approach significantly improves both translation performance and DP prediction accuracy. 

\end{abstract}

\section{Introduction}

Pronouns are important in natural languages as they imply rich discourse information. However, in pro-drop languages such as Chinese and Japanese, pronouns are frequently omitted when their referents can be pragmatically inferred from the context.
When translating sentences from a pro-drop language into a non-pro-drop language (\eg Chinese-to-English), translation models generally fail to translate invisible dropped pronouns (DPs).
This phenomenon leads to various translation problems in terms of completeness, syntax and even semantics of translations. A number of approaches have been investigated for DP translation~\cite{Nagard:2010:ACL,xiang2013enlisting,wang2016naacl,Wang:2018:AAAI}.

~\newcite{Wang:2018:AAAI} is a pioneering work to model DP translation for neural machine translation (NMT) models. They employ two {\em separate} reconstructors~\cite{Tu:2017:AAAI} to respectively reconstruct encoder and decoder representations back to the DP-annotated source sentence. The annotation of DP is provided by an {\em external} prediction model, which is trained on the parallel corpus using automatically learned alignment information~\cite{wang2016naacl}.
Although this model achieved significant improvements, there nonetheless exist two drawbacks: 1) there is no interaction between the two separate reconstructors, which misses the opportunity to exploit useful relations between encoder and decoder representations; and 2) the external DP prediction model only has an accuracy of 66\% in F1-score, which propagates numerous errors to the translation model.

In this work, we propose to improve the original model from two perspectives. First, we use a {\em shared} reconstructor to read hidden states from both encoder and decoder. Second, we integrate a DP predictor into NMT to {\em jointly} learn to translate and predict DPs. Incorporating these as two auxiliary loss terms can guide both the encoder and decoder states to learn critical information relevant to DPs.
Experimental results on a large-scale Chinese--English subtitle corpus show that the two modifications can accumulatively improve translation performance, and the best result is +1.5 BLEU points better than that reported by~\newcite{Wang:2018:AAAI}.
In addition, the jointly learned DP prediction model significantly outperforms its external counterpart by 9\% in F1-score.

\section{Background}

\begin{figure}[t]
\begin{center}
    \includegraphics[width=0.48\textwidth]{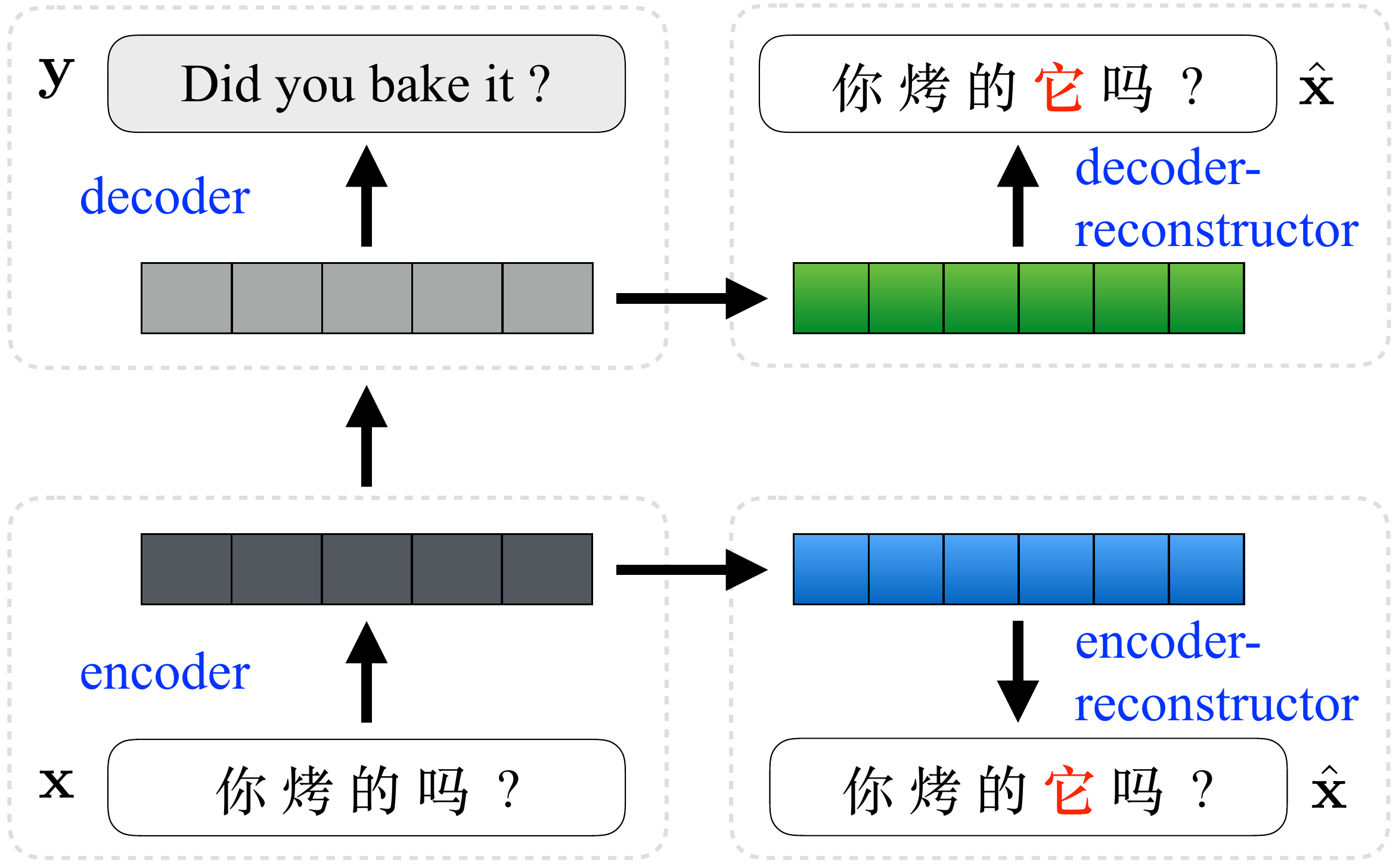}
\end{center}
\caption{Architecture of separate reconstructors.}
\label{fig:1}
\end{figure}

As shown in Figure~\ref{fig:1}, \newcite{Wang:2018:AAAI} introduced two independent reconstructors with their own parameters, which reconstruct the DP-annotated source sentence from the encoder and decoder hidden states, respectively. The central idea underpinning their approach is to guide the corresponding hidden states to embed the recalled source-side DP information and subsequently to help the NMT model generate the missing pronouns with these enhanced hidden representations.

\begin{CJK}{UTF8}{gbsn}
\begin{table}[t]
\renewcommand\arraystretch{1.1}
\centering
\begin{tabular}{l|cl}
\bf Prediction & \bf F1-score & \bf Example  \\\hline
DP Position & 88\% & {\bf 你} 烤 的 {\color{red} \#DP\#} 吗 ?\\
\hline
DP Words & 66\% & {\bf 你} 烤 的 {\bf \color{red} 它} 吗 ?\\
\end{tabular}
\caption{Evaluation of external models on predicting the positions of DPs (``DP Position'') and the exact words of DP (``DP Words'').}\label{tab:1}
\end{table}
\end{CJK}


The DPs can be automatically annotated for training and test data using two different strategies~\cite{wang2016naacl}. In the {\em training phase}, where the target sentence is available, we annotate DPs for the source sentence using alignment information.  These annotated source sentences can be used to build a neural-based DP predictor, which can be used to annotate test sentences since the target sentence is not available during the {\em testing phase}. As shown in Table~\ref{tab:1}, \newcite{wang2016naacl,Wang:2018:AAAI} explored to predict the exact DP words\footnote{Unless otherwise indicated, in the paper, the terms ``DP'' and ``DP word'' are identical.}, the accuracy of which is only 66\% in F1-score. By analyzing the translation outputs, we found that 16.2\% of errors are newly introduced and caused by errors from the DP predictor. Fortunately, the accuracy of predicting DP positions (DPPs) is much higher, which provides the chance to alleviate the error propagation problem.
Intuitively, we can learn to generate DPs at the predicted positions using a jointly trained DP predictor, which is fed with informative representations in the reconstructor. 

\section{Approach}

\begin{figure*}[h]
\begin{center}
        \subfloat[Shared reconstructor.]{
            \includegraphics[width=0.45\textwidth]{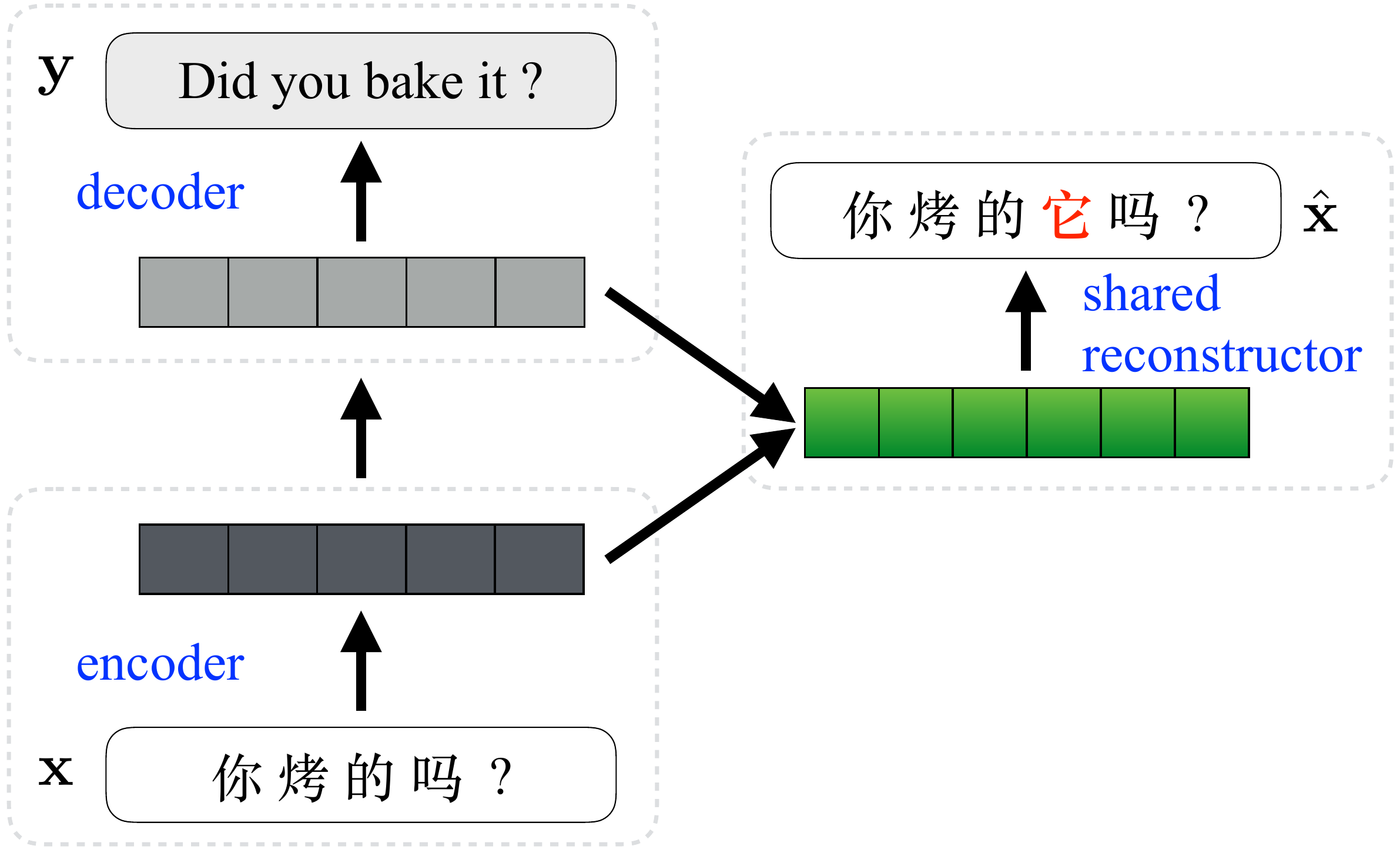}}\hfill
        \subfloat[Shared reconstructor with joint prediction.]{
            \includegraphics[width=0.45\textwidth]{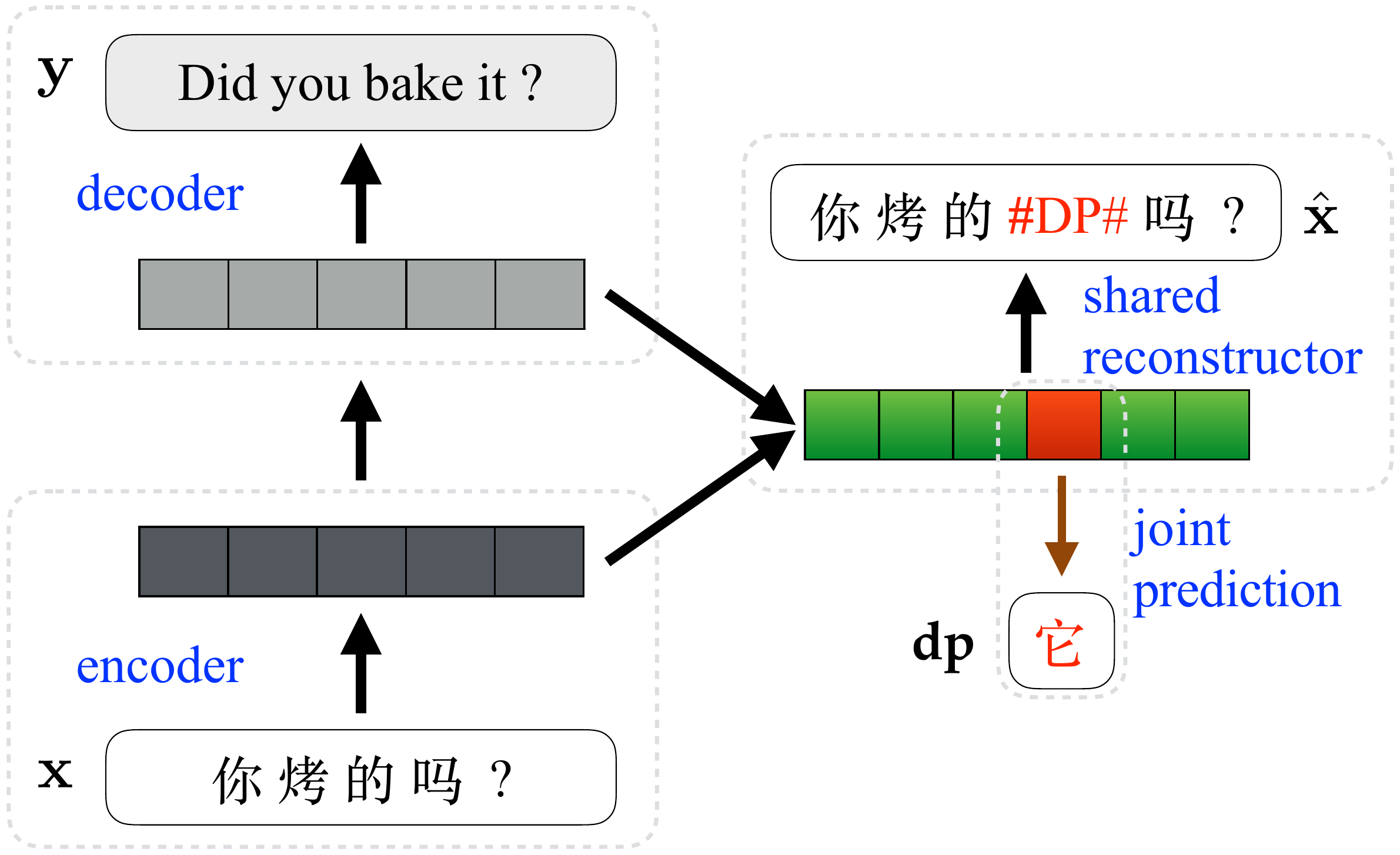}}
\end{center}
\caption{Model architectures in which the words in red are automatically annotated DPs and DPPs.}
\label{fig:2}
\end{figure*}

\subsection{Shared Reconstructor}


Recent work shows that NMT models can benefit from sharing a component across different tasks and languages. Taking multi-language translation as an example,~\newcite{Firat:2016:NAACL} share an attention model across languages while~\newcite{Dong:2015:ACL} share an encoder.
Our work is most similar to the work of~\newcite{Zoph:2016:NAACL} and~\newcite{Anastasopoulos:2018:NAACL}, which share a decoder and two separate attention models to read from two different sources. In contrast, we share information at the level of reconstructed frames.

The architectures of our proposed shared reconstruction model are shown in Figure~\ref{fig:2}(a). Formally, the reconstructor reads from both the encoder and decoder hidden states, as well as the DP-annotated source sentence, and outputs a reconstruction score. 
It uses two separate attention models to reconstruct the annotated source sentence $\hat{\bf x}=\{\hat{x}_1, \hat{x}_2, \dots, \hat{x}_T\}$ word by word, and the reconstruction score is computed by
\begin{eqnarray}
 R(\hat{\bf x}|{\bf h}^{enc}, {\bf h}^{dec})  = \prod_{t=1}^{T} g_r(\hat{x}_{t-1}, {\bf h}^{rec}_t, \hat{\bf c}^{enc}_t, \hat{\bf c}^{dec}_t) \nonumber
\end{eqnarray}
where ${\bf h}^{rec}_t$ is the hidden state in the reconstructor, and computed by Equation~(\ref{eq:1}):
\begin{eqnarray}\label{eq:1}
    {\bf h}^{rec}_t &=& f_r(\hat{x}_{t-1}, {\bf h}^{rec}_{t-1}, \hat{\bf c}^{enc}_t, \hat{\bf c}^{dec}_t)
\end{eqnarray}
Here $g_r(\cdot)$ and $f_r(\cdot)$ are respectively softmax and activation functions for the reconstructor. The context vectors $\hat{\bf c}^{enc}_t$ and $\hat{\bf c}^{dec}_t$ are the weighted sum of ${\bf h}^{enc}$ and ${\bf h}^{dec}$, respectively, as in Equation~(\ref{eq:2}) and (\ref{eq:3}):
\begin{eqnarray}
\hat{\bf c}^{enc}_t &= \sum_{j=1}^{J}{\hat{\alpha}^{enc}_{t,j}\cdot {\bf h}^{enc}_j} \label{eq:2}\\
\hat{\bf c}^{dec}_t &= \sum_{i=1}^{I}{\hat{\alpha}^{dec}_{t,i}\cdot {\bf h}^{dec}_i}\label{eq:3}
\end{eqnarray}
Note that the weights $\hat{\alpha}^{enc}$ and $\hat{\alpha}^{dec}$ are calculated by two separate attention models. We propose two attention strategies which differ as to whether the two attention models have interactions or not.

\paragraph{Independent Attention} 
calculates the two weight matrices independently, as in Equation~(\ref{eq:4}) and (\ref{eq:5}):
    \begin{eqnarray}
        \hat{\alpha}^{enc} &=& \textsc{Att}_{enc}(\hat{x}_{t-1}, {\bf h}^{rec}_{t-1}, {\bf h}^{enc}) \label{eqn:enc}\label{eq:4}\\
        \hat{\alpha}^{dec} &=& \textsc{Att}_{dec}(\hat{x}_{t-1}, {\bf h}^{rec}_{t-1}, {\bf h}^{dec}) \label{eqn:dec}\label{eq:5}
    \end{eqnarray}
where $\textsc{Att}_{enc}(\cdot)$ and $\textsc{Att}_{dec}(\cdot)$ are two separate attention models with their own parameters.

\paragraph{Interactive Attention} 
feeds the context vector produced by one attention model to another attention model. The intuition behind this is that the interaction between two attention models can lead to a better exploitation of the encoder and decoder representations. As the interactive attention is directional, we have two options (Equation~(\ref{eq:6}) and (\ref{eq:7})) which modify either $\textsc{Att}_{enc}(\cdot)$ or $\textsc{Att}_{dec}(\cdot)$ while leaving the other one unchanged:

\begin{itemize}
  \item {\em enc$\rightarrow$dec}:
    \begin{eqnarray}
    \hat{\alpha}^{dec} = \textsc{Att}_{dec}(\hat{x}_{t-1}, {\bf h}^{rec}_{t-1}, {\bf h}^{dec}, \hat{\bf c}^{enc}_t)\label{eq:6}
    \end{eqnarray}
  \item {\em dec$\rightarrow$enc}:
    \begin{eqnarray}
    \hat{\alpha}^{enc} = \textsc{Att}_{enc}(\hat{x}_{t-1}, {\bf h}^{rec}_{t-1}, {\bf h}^{enc}, \hat{\bf c}^{dec}_t)\label{eq:7}
\end{eqnarray}
\end{itemize}


\subsection{Joint Prediction of Dropped Pronouns}
\label{sec:joint}

Inspired by recent successes of multi-task learning~\cite{Dong:2015:ACL,Luong:2016:ICLR}, we propose to jointly learn to translate and predict DPs (as shown in Figure~\ref{fig:2}(b)). To ease the learning difficulty, we leverage the information of DPPs predicted by an external model, which can achieve an accuracy of 88\% in F1-score. Accordingly, we transform the original DP prediction problem to DP word generation given the pre-predicted DP positions. Since the DPP-annotated source sentence serves as the reconstructed input, we introduce an additional {\em DP-generation loss}, which measures how well the DP is generated from the corresponding hidden state in the reconstructor.

Let ${\bf dp} = \{dp_1, dp_2, \dots, dp_D\}$ be the list of DPs in the annotated source sentence, and ${\bf h}^{rec}=\{{\bf h}^{rec}_1, {\bf h}^{rec}_2, \dots, {\bf h}^{rec}_D\}$ be the corresponding hidden states in the reconstructor. The generation probability is computed by
\begin{equation}
\begin{split}
P({\bf dp}| {\bf h}^{rec}) &= \prod_{d=1}^{D} P(dp_d| {\bf h}^{rec}_{d}) \\
&= \prod_{d=1}^{D} g_{p}(dp_d| {\bf h}^{rec}_{d}) 
\end{split}
\end{equation}
where $g_p(\cdot)$ is softmax for the DP predictor.

\subsection{Training and Testing}

We train both the encoder-decoder and the shared reconstructors together in a single end-to-end process, and the training objective is
\begin{equation}
\begin{split}
J(\theta, \gamma, \psi) &= \argmax_{\theta, \gamma, \psi} \bigg\{ \underbrace{\log L({\bf y}|{\bf x}; \theta)}_\text{\normalsize \em likelihood} \\
&+ \underbrace{\log R({\bf \hat{x}} | {\bf h}^{enc}, {\bf h}^{dec}; \theta, \gamma)}_\text{\normalsize \em reconstruction} \\
&+ \underbrace{\log P({\bf dp} | \hat{\bf h}^{rec}; \theta, \gamma, \psi)}_\text{\normalsize \em prediction} \bigg\}
\end{split}
\end{equation}
where $\{\theta, \gamma, \psi\}$ are respectively the parameters associated with the encoder-decoder, shared reconstructor and the DP prediction model. 
The auxiliary reconstruction objective $R(\cdot)$ guides the related part of the parameter matrix $\theta$ to learn better latent representations, which are used to reconstruct the  DPP-annotated source sentence. 
The auxiliary prediction loss $P(\cdot)$ guides the related part of both the encoder-decoder and the reconstructor to learn better latent representations, which are used to predict the DPs in the source sentence.

\begin{table*}[t]
\centering
\begin{tabular}{c|l|r|cc|l}
	\multirow{2}{*}{\bf \#} &   \multirow{2}{*}{\bf Model}        & \multirow{2}{*}{\bf \#Params}  &  \multicolumn{2}{c|}{\bf Speed}  &   \multirow{2}{*}{\bf BLEU}\\
	\cline{4-5}
	   &  &  & Train    &   Decode    &  \\
	\hline\hline
	\multicolumn{6}{c}{Existing system \cite{Wang:2018:AAAI}}\\
	\hline
	1   &   Baseline                &  86.7M &   1.60K   &  {15.23}    & 31.80\\   
	2   &   Baseline (+DPs)         &  86.7M  &   1.59K    &   {15.20}   & 32.67\\
	\hline
	3   &   Separate-Recs$\Rightarrow$(+DPs)  & +73.8M  &   0.57K   &   {12.00}   & 35.08\\
	\hline\hline
	\multicolumn{6}{c}{Our system} \\
	\hline
	4   &   Baseline (+DPPs)        &  86.7M  &   1.54K    &   {15.19}   &    33.18\\
    \hline
	5   &   Shared-Rec$_{independent}$$\Rightarrow$(+DPPs) & +86.6M & 0.52K & {11.87} & 35.27$^{\dag\ddag}$\\
	6   &   Shared-Rec$_{independent}$$\Rightarrow$(+DPPs) + joint prediction & +87.9M & 0.51K & {11.88} & 35.88$^{\dag\ddag}$\\
    7   &   	Shared-Rec$_{enc \rightarrow dec}$$\Rightarrow$(+DPPs) + joint prediction & +91.9M & 0.48K & {11.84} & \bf 36.53$^{\dag\ddag}$\\
	8   &   Shared-Rec$_{dec \rightarrow enc}$$\Rightarrow$(+DPPs) + joint prediction & +89.9M & 0.49K & {11.85} & 35.99$^{\dag\ddag}$
\end{tabular}
\caption{\label{tab:2} Evaluation of translation performance for Chinese--English. ``Baseline'' is trained and evaluated on the original data, while ``Baseline (+DPs)'' and ``Baseline (+DPPs)'' are trained on the data annotated with DPs and DPPs, respectively. Training and decoding (beam size is 10) speeds are measured in words/second. ``$\dag$'' and ``$\ddag$'' indicate statistically significant difference ($p < 0.01$) from ``Baseline (+DDPs)'' and ``Separate-Recs$\Rightarrow$(+DPs)'', respectively.}\label{tab-results}
\end{table*}

Following~\citeauthor{Tu:2017:AAAI}~\shortcite{Tu:2017:AAAI} and \citeauthor{Wang:2018:AAAI}~\shortcite{Wang:2018:AAAI}, we use the reconstruction score as a reranking technique to select the best translation candidate from the generated $n$-best list at testing time. Different from~\citeauthor{Wang:2018:AAAI}~\shortcite{Wang:2018:AAAI}, we reconstruct DPP-annotated source sentence, which is predicted by an external model.



\section{Experiment}

\subsection{Setup}

To compare our work with the results reported by previous work~\cite{Wang:2018:AAAI}, we conducted experiments on their released Chinese$\Rightarrow$English TV Subtitle corpus.\footnote{\url{https://github.com/longyuewangdcu/tvsub}}
The training, validation, and test sets contain 2.15M, 1.09K, and 1.15K sentence pairs, respectively. 
We used case-insensitive 4-gram NIST BLEU metrics \cite{Papineni:2002} for evaluation, and {\em sign-test} \cite{Collins05} to test for statistical significance.

We implemented our models on the code repository released by~\newcite{Wang:2018:AAAI}.\footnote{\url{https://github.com/tuzhaopeng/nmt}} We used the same configurations (\eg vocabulary size = 30K, hidden size = 1000) and reproduced their reported results. It should be emphasized that we did not use the pre-train strategy as done in~\newcite{Wang:2018:AAAI}, since we found training from scratch achieved a better performance in the shared reconstructor setting.

\subsection{Results}

Table~\ref{tab-results} shows the translation results.
It is clear that the proposed models significantly outperform the baselines in all cases, although there are considerable differences among different variations.

\paragraph{Baselines} (Rows 1-4):
The three baselines (Rows 1, 2, and 4) differ regarding the training data used. 
``Separate-Recs$\Rightarrow$(+DPs)'' (Row 3) is the best model reported in~\newcite{Wang:2018:AAAI}, which we employed as another strong baseline.
The baseline trained on the DPP-annotated data (``Baseline (+DPPs)'', Row 4) outperforms the other two counterparts, indicating that the error propagation problem does affect the performance of translating DPs. It suggests the necessity of jointly learning to translate and predict DPs.

\paragraph{Our Models} (Rows 5-8):
Using our shared reconstructor (Row 5) not only outperforms the corresponding baseline (Row 4), but also surpasses its separate reconstructor counterpart (Row 3). Introducing a joint prediction objective (Row 6) can achieve a further improvement of +0.61 BLEU points. These results verify that shared reconstructor and jointly predicting DPs can accumulatively improve translation performance.

Among the variations of shared reconstructors (Rows 6-8), we found that an interaction attention from encoder to decoder (Row 7) achieves the best performance, which is +3.45 BLEU points better than our baseline (Row 4) and +1.45 BLEU points better than the best result reported by~\newcite{Wang:2018:AAAI} (Row 3). We attribute the superior performance of ``Shared-Rec$_{enc\rightarrow dec}$'' to the fact that the attention context over encoder representations embeds useful DP information, which can help to better attend to the representations of the corresponding pronouns in the decoder side. 
Similar to~\citeauthor{Wang:2018:AAAI}~\shortcite{Wang:2018:AAAI}, the proposed approach improves BLEU scores at the cost of decreased training and decoding speed, which is due to the large number of newly introduced parameters resulting from the incorporation of reconstructors into the NMT model.

\subsection{Analysis}

\begin{table}[h]
\renewcommand\arraystretch{1.1}
\centering
\begin{tabular}{l|ccc}
\bf Models & \bf Precision & \bf Recall & \bf F1-score  \\\hline
External & 0.67 & 0.65 & 0.66 \\\hline
Joint & 0.74 & 0.76 & 0.75 \\
\end{tabular}
\caption{Evaluation of DP prediction accuracy. ``External'' model is {\em separately} trained on DP-annotated data with external neural methods \cite{wang2016naacl}, while ``Joint'' model is {\em jointly} trained with the NMT model (Section~\ref{sec:joint}). }\label{tab:3}
\end{table}

\paragraph{DP Prediction Accuracy} 
As shown in Table~\ref{tab:3},
the jointly learned model significantly outperforms the external one by 9\% in F1-score. We attribute this to the useful contextual information embedded in the reconstructor representations, which are used to generate the exact DP words.

\begin{table}[h]
\centering
\renewcommand\arraystretch{1.1}
\begin{tabular}{l|c|c}
	\bf Model & \bf Test & \bf $\bigtriangleup$ \\
	\hline
	Baseline (+DPPs)     &   33.18   &   -- \\
	Separate-Recs (+DPs)           &   34.02   &   +0.84\\
	\hline
    Shared-Rec (+DPPs) & {\bf 34.80}   &   {\bf +1.62}\\
\end{tabular}
\caption{\label{tab-results-no-rec}  Translation results when {\em reconstruction is used in training only while not used in testing}.}
\end{table}

\paragraph{Contribution Analysis}

Table~\ref{tab-results-no-rec} lists translation results when the reconstruction model is used in training only. We can see that the proposed model outperforms both the strong baseline and the best model reported in~\newcite{Wang:2018:AAAI}. This is encouraging since no extra resources and computation are introduced to online decoding, which makes the approach highly practical, for example for translation in industry applications.

\begin{table}[h]
\centering
\renewcommand\arraystretch{1.1}
\begin{tabular}{l|c|c|c}
	\bf Model & \bf Auto.   &   \bf Man.   & \bf $\bigtriangleup$ \\
	\hline
    Seperate-Recs (+DPs)   & 35.08 &   38.38   &   +3.30\\
	Shared-Rec (+DPPs)   & 36.53 &   38.94   &   +2.41
\end{tabular}
\caption{\label{tab-results-DP-accuracy} Translation performance gap (``$\bigtriangleup$'') between manually (``Man.'') and automatically (``Auto.'') labelling DPs/DPPs for input sentences in testing.}
\end{table}

\paragraph{Effect of DPP Labelling Accuracy}

For each sentence in testing, the DPs and DPPs are labelled automatically by two separate external prediction models, the accuracy of which are respectively 66\% and 88\% measured in F1 score. We investigate the best performance the models can achieve with manual labelling, which can be regarded as an ``Oracle'', as shown in Table~\ref{tab-results-DP-accuracy}. As seen, there still exists a significant gap in performance, and this could be improved by improving the accuracy of our DPP generator. In addition, our models show a relatively smaller distance in performance from the oracle performance (``Man''), indicating that the error propagation problem is alleviated to some extent.

\section{Conclusion}

In this paper, we proposed effective approaches of translating DPs with NMT models: {\em shared} reconstructor and {\em jointly} learning to translate and predict DPs. Through experiments we verified that 1) shared reconstruction is helpful to share knowledge between the encoder and decoder; and 2) joint learning of the DP prediction model indeed alleviates the error propagation problem by improving prediction accuracy. The two approaches accumulatively improve translation performance. The method is not restricted to the DP translation task and could potentially be applied to other sequence generation problems where additional source-side information could be incorporated.

In future work we plan to: 1) build a fully end-to-end NMT model for DP translation, which does not depend on any external component (\ie DPP predictor); 2) exploit cross-sentence context \cite{wang2017exploiting} to further improve DP translation; 3) investigate a new research strand that adapts our model in an inverse translation direction by learning to drop pronouns instead of recovering DPs.

\section*{Acknowledgments}
The ADAPT Centre for Digital Content Technology is funded under the SFI Research Centres Programme (Grant 13/RC/2106) and is co-funded under the European Regional Development Fund. 
We thank the anonymous reviewers for their insightful comments.

\balance
\bibliographystyle{acl_natbib_nourl}
\bibliography{emnlp2018.bib}
\end{document}